# Novel Magnetic Actuation Strategies for Precise Ferrofluid Marble Manipulation in Magnetic Digital Microfluidics: Position Control and Applications

Mohammad Hossein Sarkhosh[1], Mohammad Hassan Dabirzadeh[1], Mohamad Ali Bijarchi[2**], and Hossein Nejat Pishkenari[1*]

[1] Micro/Nano Robotics Laboratory, Department of Mechanical Engineering, Sharif University of Technology, Tehran, Iran

[2] Department of Mechanical Engineering, Sharif University of Technology, Tehran, Iran

[*] Corresponding author: nejat@sharif.edu

[**] Corresponding author: bijarchi@sharif.edu

**Abstract**

Precise manipulation of liquid marbles has significant potential in various applications such as lab-on-a-chip systems, drug delivery, and biotechnology and has been a challenge for researchers. Ferrofluid marble (FM) is a marble with a ferrofluid core that can easily be manipulated by a magnetic field. Although FMs have great potential for accurate positioning and manipulation, these marbles have not been precisely controlled in magnetic digital microfluidics, so far. In this study for the first time, a novel method of magnetic actuation is proposed using a pair of Helmholtz coils and permanent magnets. The governing equations for controlling the FM position are investigated, and it is shown that there are three different strategies for adjusting the applied magnetic force. Then, experiments are conducted to demonstrate the capability of the proposed method. To this aim, different magnetic setups are proposed for manipulating FMs. These setups are compared in terms of energy consumption and tracking ability across various frequencies. The study showcases several applications of precise FM position control, including controllable reciprocal positioning, simultaneous position control of two FMs, the transport of non-magnetic liquid marbles using the FMs, and sample extraction method from the liquid core of the FM.

# 1.Introduction

Liquid handling in microfluidics systems falls into two main categories: continuous-flow microfluidics and droplet-based microfluidics, each offering distinct advantages. Droplet-based microfluidics, due to its ability to handle discrete fluid components, has been applied in areas such as pharmaceutical research, microparticle preparation, cell analysis, and nutrition by enabling the generation, manipulation, and control of droplets [1–6]. To achieve more precise control over individual fluid components, digital microfluidics has emerged, allowing for manipulating small fluid volumes—such as droplets or liquid marbles—while automating complex laboratory processes. This approach reduces the amount of reagents and samples needed for experiments, making it more efficient [7].

Liquid marble emerges as a highly promising and distinctive digital microfluidic platform capable of effortlessly merging , splitting [8,9], deforming [10], mixing [11], and transporting [12]. A liquid marble is a type of liquid-solid material that consists of a liquid droplet encapsulated by a layer of hydrophobic powder [13,14]. The hydrophobic layer prevents the liquid from spreading out and wetting the surface, giving it the appearance and behavior of a solid marble [15]. This characteristic offers significant advantages for improving and diversifying the means of transporting small liquids, extending its applicability to various microfluidic applications [8,16]. Manipulating the marbles can be achieved by activating them through optical [17], electrowetting [18,19], magnetic [20], and acoustic [21] manipulations. Among these methods, magnetic actuation stands out as an accessible, easy-to-use, wireless, and cost-effective approach, utilizing either electromagnets or permanent magnets [22]. Creating magnetic properties involves integrating iron particles within either the core or the shell of the marble. Typically, ferrofluid is employed within the core, resulting in what is known as FM.

Various magnetic actuation techniques are employed in magnetic digital microfluidics, including magnetic particle-based manipulation [22], permanent magnet-assisted manipulation [23], and ferrofluid-based magnetic droplet manipulation [22]. Extensive research has been conducted on ferrofluids due to their unique properties, simplifying manipulation. Ferrofluid Consists of magnetic nanoparticles suspended in a carrier fluid, behaving like regular liquids while responding to magnetic fields like solids. It is a remarkable material that has captivated the attention of scientists and engineers for its exceptional properties and diverse applications, originally developed in the 1960s for space applications [24–28]. Magnetic stimuli create capability for ferrofluid transporting [26,29,30], merging [31–33], splitting [31,33–35], droplet generating [36,37], and shape control [33,38], using a permanent magnet or an electromagnet. As a lab-on-chip platform, Lin et al. developed a magnetic digital microfluidic setup for overcoming the SARS-Covid-19 pandemic. They employed an array of printed circuit board (PCB) electromagnets in conjunction with millimetric permanent magnets. This system effectively manipulated the ferrofluid droplets. Through this method, their approach seamlessly integrated merging, mixing, and dispensing processes, culminating in the successful development of a platform for nucleic acid amplification [39].

Controlling the position of microrobots using a magnetic field has abroad application in biomedical surgery, drug delivery, and chemistry [40]. Several researchers have introduced micro-scale solid permanent magnets serving as microrobots and proposed multiple techniques to control their position. For instance, controlling the position of the multiple magnetic microrobots independently in a 2D workspace [41,42] or a single magnetic microrobot in a 3D workspace using a reinforcement learning algorithm [43]. None of these microrobots offer the same level of flexibility as provided by ferrofluid soft robots. This flexibility encompasses capabilities such as splitting, merging, and deforming. For instance, Ahmed et al. utilized a setup consisting of eight electromagnetic coils and a PID controller to control the length and movement of a ferrofluid droplet [38]. They also demonstrated the ability to move microparticles using ferrofluidic robots. Fan et al. utilized ferrofluid droplets as a miniature soft robot, offering programmable deformations for versatile locomotion and manipulation. Unlike conventional elastomer-based robots, these liquid droplets can be navigated through highly constrained environments and exhibit various functionalities, such as on-demand splitting, merging, and cooperative tasks [44]. Fan et al. focused on scalable ferrofluidic robots, proposing a setup that uses electromagnets and a permanent magnet. This innovative approach allows for scalability across different sizes of ferrofluidic robots, ranging from centimeters to micrometers, thereby expanding the capabilities of ferrofluidic robotics [45].

Magnetic liquid marbles have gained attention as miniature soft machines, thanks to their magnetic properties, with practical applications in sensors, bioreactors, and microfluidic systems [46–48]. Their wireless control and soft-bodied nature make them ideal for liquid handling technologies and precise operations in magnetofluidics [48]. Bormashenko et al. created FM by combining polyvinylidene fluoride and $\gamma Fe_3O_4$ powder to coat a ferrofluid droplet. They successfully showcased that the FM could attain a velocity of 25 cm/s under a magnetic flux density of 0.5 mT generated by a permanent magnet [49]. Previous manipulation methods for FMs have been limited in accuracy and functionality; Sarkhosh et al. manipulated ferrofluid marble by a repulsive force for the first time using Helmholtz coils and permanent magnets [29]. Dayyani et al. explored floating liquid marbles with magnetic shells and deionized water cores, actuated by varying magnetic fields [50]. Khaw et al. placed a permanent magnet on a programmable linear stage to locate liquid marble with Magnetite particles inside of it [51].

Gaining precise control over the position of liquid marbles remains a significant challenge in digital microfluidics. While various magnetic actuation methods have been explored in previous studies, the absence of precise position steering for FMs is notable. Considering the great potential of the FMs by the magnetic field for precise manipulation, this research introduces, for the first time, the use of repulsive magnetic force to develop three distinct methods for position steering of the FMs. This novel approach integrates a combination of Helmholtz coils and permanent magnets to exert precise control over FMs. Furthermore, the study outlines the capabilities of these setups and proposes a comprehensive and fair comparison between them. Finally, potential applications for precise position control of FMs are discussed, including simultaneous control of multiple FMs, mixing and transporting capabilities, carrying non-magnetic marbles, and sample extraction, highlighting their versatility in advanced microfluidic systems.

## 2. Materials and methods

### *2.1. Theoretical formulas and scenarios of position control*

### *2.1.1. Setup 1: Position control of the FM using the variable Helmholtz coils current*

The repulsive magnetic force is produced when similar magnetic poles are facing each other. The FMs do not inherently possess a magnetic dipole moment in the absence of an external magnetic field. When the FM is placed in a uniform magnetic field, the magnetic dipole moment is induced without experiencing a net force. When a permanent magnet is placed near an induced FM with a similar dipole, the FM is manipulated by a repulsive magnetic force. Figure 1(a) schematically illustrates this setup, where Helmholtz coils generate a uniform magnetic field and a permanent magnet is employed to repel the FM. In this setup, the motion of the FM is constrained by the walls of a water pool allowing movement along the x-direction. Additionally, the FM floats on the water to minimize the friction force. The camera shown in Figure 1(a), localizes the FM in the x-direction.

Figure 1(b) illustrates the various reactions of the FM under different currents applied to the Helmholtz coils. As shown, when the current in the Helmholtz coils is positive and the dipole moment of the permanent magnet is similar to that of induced FM, a repulsive magnetic force is exerted on the FM. As the current in the Helmholtz coils is reduced, the induced magnetic dipole in the FM weakens, resulting in a decrease in the repulsive magnetic force. When the current of Helmholtz coils is inactivated, the magnetic field of the permanent magnet induces an opposite dipole direction in the FM, leading to production an attractive force. When the current in the Helmholtz coils is reversed, a stronger magnetic dipole moment is induced in the FM, leading to a stronger attractive magnetic force. Hence, the FM position could be controlled by tuning the magnitude and direction of current in Helmholtz coils. In fact, it gets away from or approaches the permanent magnet in response to the repulsive or attractive magnetic forces, respectively.

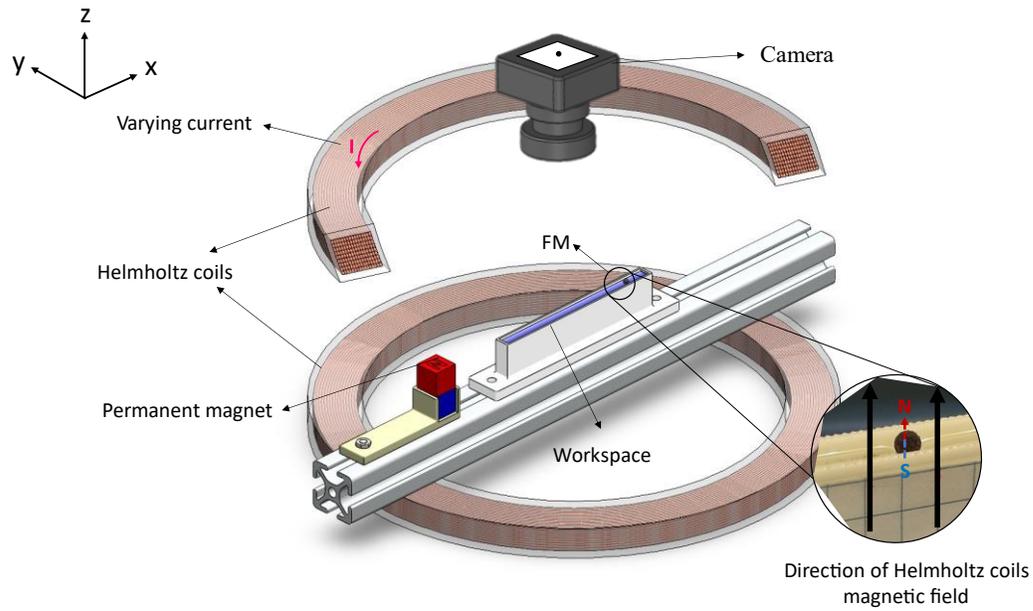

(a)

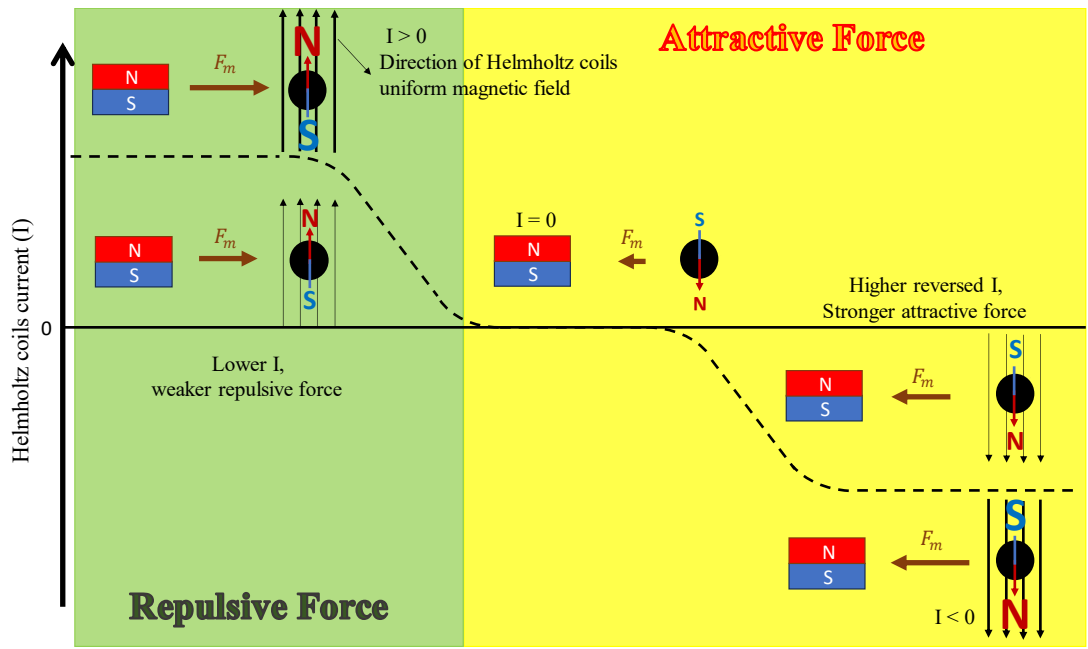

(b)

Figure 1: (a) Schematic of Setup 1 which benefits from a fixed permanent magnet and variable Helmholtz coils current, (b) various response of the FM in different current of the Helmholtz coils.

The dynamic response of the FM on the water surface is proposed in Eq (1) in the x-direction of movement [29]:

$$F_m - F_f = m\ddot{x} \tag{1}$$

where $F_m$ is the magnetic force, $m$ is mass of the FM, $F_f$ is the friction force exerted to the FM, as written as Eq (2):

$$F_f = 6\pi\beta R\mu\dot{x} \tag{2}$$

where $R$ is the FM radius, $\mu$ is the dynamic viscosity of the FM, and $\beta$ is the correction factor for the friction coefficient [52]. Also, $\boldsymbol{F_m}$ can be evaluated from Eq (3):

$$\boldsymbol{F_m} = \frac{\chi V}{\mu_0} \boldsymbol{B}\nabla\boldsymbol{B} \tag{3}$$

where $\chi$ is the ferrofluid magnetic susceptibility, $\mu_0$ is the magnetic permeability of the vacuum, and $B$ is the total magnetic flux density (induced by the Helmholtz coils and the permanent magnet). The gradient of the magnetic flux density which is produced by the Helmholtz coils is assumed to be zero at the workspace. So, the only source of the magnetic flux density gradient is from the permanent magnet ($\nabla \boldsymbol{B}_{PM}$) which can be achieved by Eq (4).

$$\nabla \boldsymbol{B} = \nabla \boldsymbol{B}_{PM} = \frac{3\mu_0|\boldsymbol{m}_{PM}|}{4\pi|\boldsymbol{r}|^4}(\widehat{\boldsymbol{m}}_{PM}\hat{\boldsymbol{r}}^T + \hat{\boldsymbol{r}}\widehat{\boldsymbol{m}}_{PM}^T - [5\hat{\boldsymbol{r}}\hat{\boldsymbol{r}}^T - \boldsymbol{U}](\widehat{\boldsymbol{m}}_{PM}.\hat{\boldsymbol{r}})) \tag{4}$$

In Eq (4), $\boldsymbol{m_{PM}}$ is the magnetization of the permanent magnet, $\widehat{\boldsymbol{m}}_{PM}$ is the unit vector of $\boldsymbol{m_{PM}}$, $\mathbf{r}$ is the vector from the center of the permanent magnet to the FM position, $\hat{\boldsymbol{r}}$ is the unit vector of $\mathbf{r}$, and $\boldsymbol{U}$ is the 3×3 identity matrix.

In this setup, the magnetic flux density is generated just in z-direction by the Helmholtz coils and the permanent magnet at the workspace. Therefore, the magnetic flux density is assumed to be as Eq (5) and magnetic flux density of the permanent magnet in z-direction is calculated from Eq (6).

$$\boldsymbol{B} = B_{Helmholtz} \begin{Bmatrix} 0 \\ 0 \\ 1 \end{Bmatrix} + B_{PM} \begin{Bmatrix} 0 \\ 0 \\ 1 \end{Bmatrix} \quad (5)$$

$$\boldsymbol{B}_{PM} = \frac{\mu_0 |\boldsymbol{m}_{PM}|}{4\pi |r|^3} (3\,\hat{r}\hat{r}^T - \boldsymbol{U}) \quad (6)$$

By assuming that the FM movement is restricted in the x-direction ($\hat{r} = [1\ 0\ 0]^T$), and the magnetic dipole of the FM is rotated about the x-direction ($\hat{\boldsymbol{m}}_{PM} = [0\ \hat{m}_{PM}^y\ \hat{m}_{PM}^z]$). By replacing $\boldsymbol{B}$ and $\nabla \boldsymbol{B}$ into Eq (3), the magnetic force can be rewritten as shown in Eq (7):

$$F_m = \pm \frac{3\mu_0\,\chi\,(B_{Helmholtz} + B_{PM})|\boldsymbol{m}_{PM}|}{4\,\pi\,r^4}\,\hat{m}_{PM}^z \quad (7)$$

where $r$ is equal to $(x_{FM} - x_{pm})$, and $\hat{m}_{pm}^z$ is the z-component of $\hat{\boldsymbol{m}}_{PM}$. According to Eq (7), two parameters can affect the amount of $F_m$: $\hat{m}_{PM}^z$ and $\pm|B_{Helmholtz}|$. The $\hat{m}_{PM}^z$ is varied when the permanent magnet is rotated, and the $\pm|B_{Helmholtz}|$ is changed by controlling the current of the Helmholtz coils. By fixing each one and varying the another, two scenarios can be proposed. In this setup (Setup 1), the Helmholtz coil current is adjusting and direction of the permanent magnet is fixed. Conversely, in the following setup (Setup 2), the permanent magnet is rotating and the current of the Helmholtz coils is held constant.

In setup1, where the magnetic dipole of a permanent magnet along the z-direction ($m_{pm}^z = 1$), the magnetic flux density which is produced by the Helmholtz coils is expressed in terms of the desired magnetic force. Consequently, Eq (7) can be reformulated as Eq (8):

$$B_{Helmholtz} = \pm \frac{F_{des}\, 4\pi r^4}{3\mu_0 \chi\, |\boldsymbol{m_{pm}}|} - \frac{\mu_0 |\boldsymbol{m_{pm}}|}{2\pi r^3} \tag{8}$$

where $B_{Helmholtz} = f(I)$, a function that will be derived for the Helmholtz coils in the "Experimental setup" section. Moreover, $I$ represents the current in the Helmholtz coils, which can act as the control input. In other words, the desired force can be generated by computing the appropriate value of $I$.

### 2.1.2. Setup 2: Position control of the FM using a single rotating permanent magnet

The schematic of Setup 2 is shown in Figure 2(a), in which the current of the Helmholtz coils is fixed, while it benefits from a rotating permanent magnet for adjusting the magnetic force. As mentioned in Eq (7), $m_{pm}^z$ is another parameter that affects the magnetic force magnitude. This parameter can be adjusted by rotating the permanent magnet. In Eq (9), $\hat{m}_{pm}^z$ is related to the angle of the permanent magnet:

$$\cos\theta = \hat{m}_{pm}^z \tag{9}$$

where $\theta$ is the angle between $\hat{\boldsymbol{m}}_{pm}$ and z-direction, so by varying the $\theta$, different value of magnetic force is exerted to the FM.

As shown in Figure 2(b), when the magnetic dipole moments of the permanent magnet and the FM are in the same direction ($\theta = 0°$), a strong repulsive force is exerted to the FM. By increasing $\theta$, the magnetic force applied to the FM reduces until $\theta = 90°$ ($\cos 90° = \hat{m}_{pm}^z = 0$) and the magnetic force reaches zero. On the other hand, when the angle of the permanent magnet is between $90° < \theta < 180°$, the attractive force is applied to the FM because its dipole moment is in the opposite direction relative to that of the FM. The greater angle of the permanent magnet relative to 90°, the more attractive magnetic force is applied to the FM. Finally, at $\theta = 180°$ the maximum amount of attractive force is exerted to the FM.

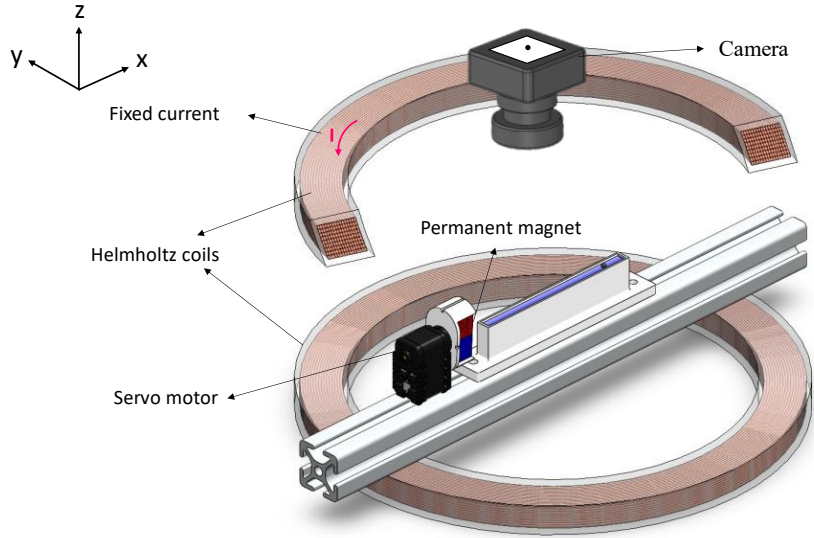

(a)

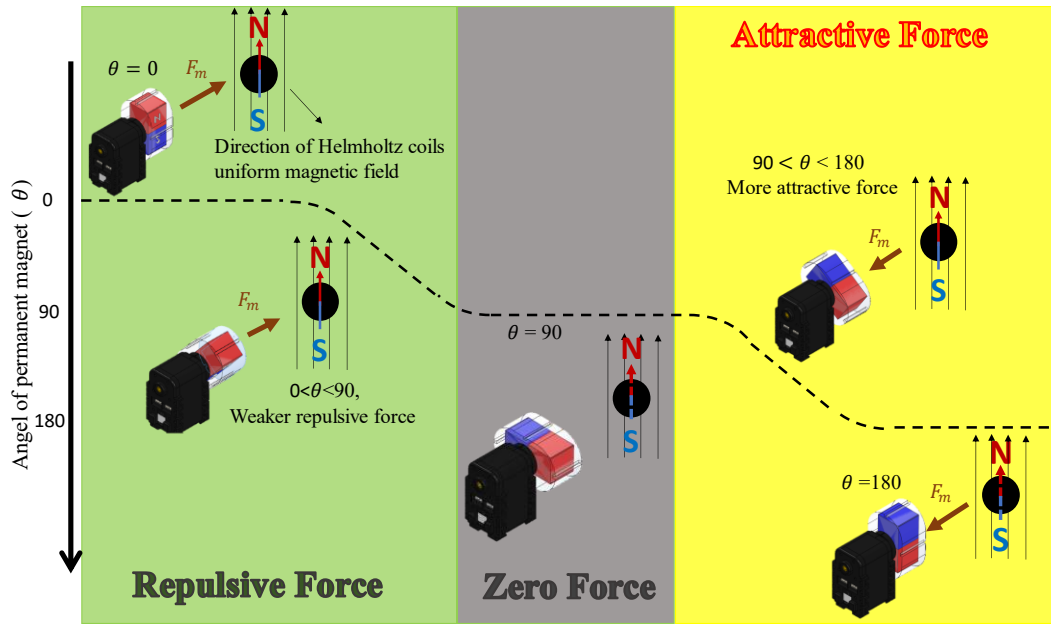

(b)

Figure 2: (a) The schematic of the Setup 2 includes a rotating permanent magnet and a constant current in the Helmholtz coils, (b) the magnetic forces applied to the FM at different angles of the permanent magnet

The direction of the magnetic flux density produced by the rotating permanent magnet continuously changes in the z- and y-directions and is negligible compared to that of the Helmholtz coils. Hence, the equations can be simplified by assuming that the magnetic flux density produced by the Helmholtz coils dominates over that of the permanent magnet ($B = B_{Helmholtz}$). Therefore, $\cos \theta$ can be rewritten in terms of the desired force as shown in Eq (10).

$$\cos \theta = \frac{F_{des} 4 \pi r^4}{3\mu_0 \chi |B_{Helmholtz}||m_{pm}|} \tag{10}$$

Therefore, in this setup, the desired force is derived in terms of the angle of the permanent magnet. Thus, the control input is the angle of the permanent magnet.

### *2.1.3. Setup 3: Position control of one FM using two rotating permanent magnets*

Setup 3 utilizes two rotating permanent magnets in each side of the workspace instead of one which is schematically depicted in Figure 3(a). The size of these two permanent magnets is half of the previous one for fair comparison. Contrary to the previous setup which utilizes attractive and repulsive magnetic forces, this setup is proposed to utilize solely repulsive magnetic force for controlling the FM position. Moreover, this setup has two degrees of freedom by using two rotating permanent magnets. Figure 3(b) depicts different resultant magnetic forces versus angles of two permanent magnets. When the angle of the permanent magnet number 1 is zero ($\theta_1 = 0°$), and the permanent magnet number 2 is ($\theta_2 = 90°$), the maximum repulsive force magnitude is applied to the FM in the negative direction of the x-axis. By decreasing the angle of the permanent magnet number 2 ($0 < \theta_2 < 90°$), the resultant magnetic force decreases. While $\theta_1 = \theta_2 = 0°$, the resultant force is zero at the center of the workspace. In order to exert the magnetic force to the FM in the positive direction of x-axis, the angle of permanent magnet number 2 is fixed at $\theta_2 = 0°$, while the angle of permanent magnet number 1 is varied between $\theta_1 = 0$ and $\theta_1 = 90$.

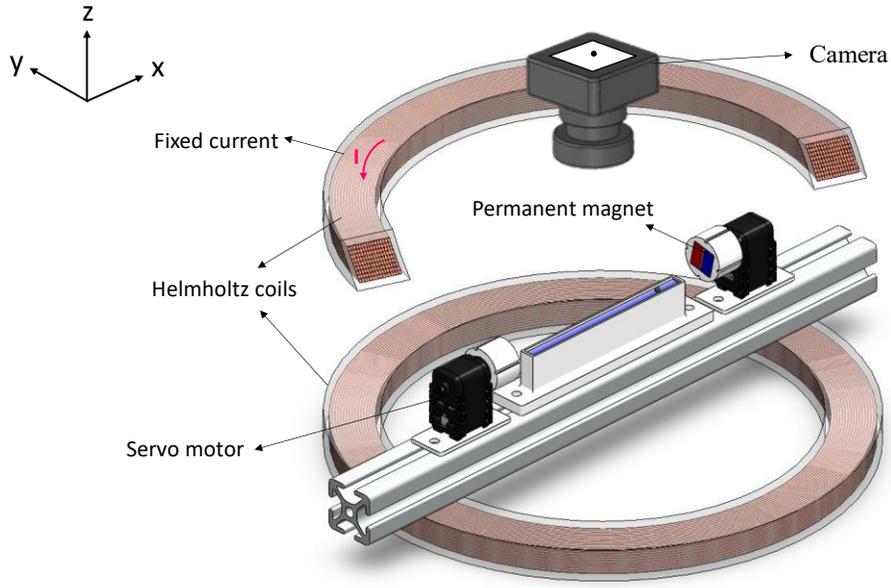

(a)

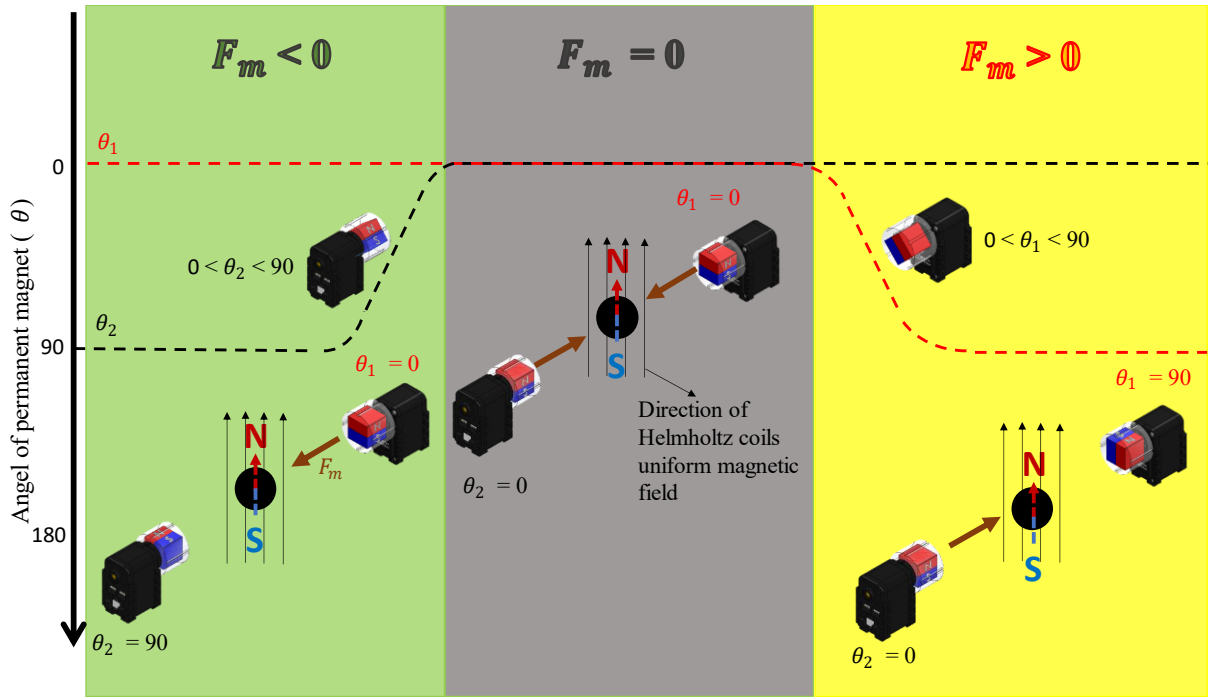

(b)

*Figure 3: (a) Schematic of the Setup 3 uses two rotating permanent magnets to control the position of the FM using only repulsive magnetic forces, (b) various reactions of FM under different angles of the two permanent magnets.*

The total force exerted on the FM by the two permanent magnets can be determined by summing the individual forces, as shown in Eq (11):

$$F_m(x) = \frac{3\mu_0\chi|B_{Helmholtz}||m_{pm}|}{4\pi r_1^4}\cos\theta_1 - \frac{3\mu_0\chi|B_{Helmholtz}||m_{pm}|}{4\pi r_2^4}\cos\theta_2 \tag{11}$$

$$= A\left(\frac{\cos\theta_1}{r_1^4} - \frac{\cos\theta_2}{r_2^4}\right)$$

where A is $\frac{3\mu_0|B_{Helmholtz}||m_{pm}|}{4\pi}$. In order to achieve the angles of two permanent magnets to exert the desired force ($F_{des}$) to the FM, Eq (11) which involves two unknown parameters, is solved. To utilize maximum repulsive magnetic force for controlling the FM position, one of the permanent magnets is fixed at angle of zero (For instance, $\theta_1 = 0$) based on the direction of the desired force ($F_d$), and the angle of another permanent magnet ($\theta_2$) is adjusted to applied the desired force. The angle of the permanent magnets are achieved form Eq (12):

$$\theta_1 = 0, \cos\theta_2 = \frac{\left(\frac{1}{r_2^4} - \frac{F_d}{A}\right)}{\frac{1}{r_1^4}} \Big/ \theta_2 = 0, \cos\theta_1 = \frac{\left(\frac{F_d}{A} - \frac{1}{r_1^4}\right)}{\frac{1}{r_2^4}} \tag{12}$$

where $\theta_1$ and $\theta_2$ are the angle of the permanent magnets. Therefore, the desired force is calculated based on the cosines of the angles of the two permanent magnets.

The details of the experimental setups are provided in Supplementary Note 1. Additionally, the magnetic flux density on the workspace, denoted as $B_{Helmholtz}$, is measured. with a precise gauss meter (TES-3196/3197) for different input currents ($I$). By fitting a linear function over the data ($B$ versus $I$) with $R^2$ equal to 0.9771, Eq (13) is obtained as follows.

$$B_{Helmholtz} = 1.817 \times I(A) \tag{13}$$

## 3. Results and discussion

The block diagram of the FM position control scheme is illustrated in Figure 4. The process begins by calculating the position error, which is the difference between the desired position ($x_d$) and the current FM position ($x$). It is worth mentioning that the FM position is calculated through one-dimensional image processing. The controller then determines the desired force to be applied to the FM to correct this error. Depending on the setup in use, this desired force is converted into the current for the Helmholtz coils (Setup 1) or the angle adjustment for the permanent magnets (Setup 2, and 3). Subsequently, the FM's position is measured again and this new position is used to recalculate the error, and the cycle repeats until the FM position aligns with the desired position.

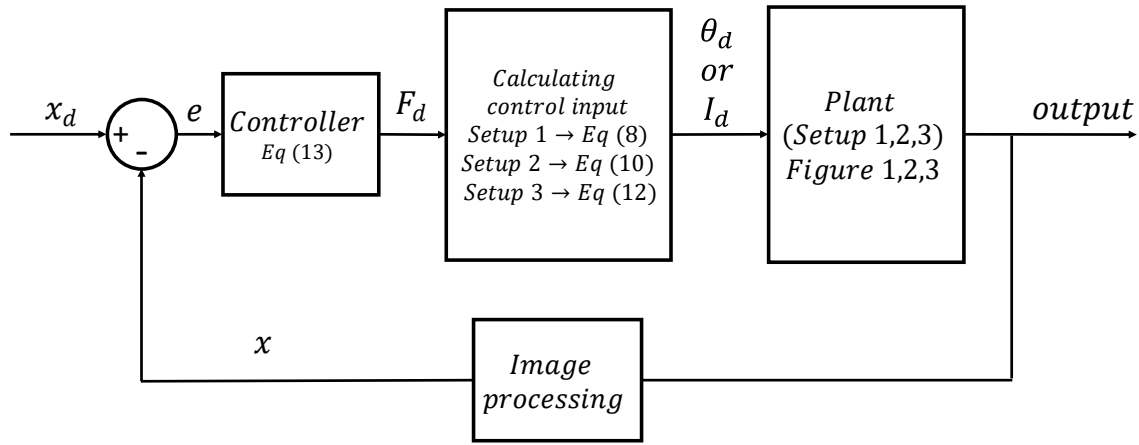

*Figure 4: block diagram of the control scheme*

In this study, a Proportional-integral (PI) controller beside a feed-forward controller is utilized for the position control of the FM. Similar controllers have been also employed by researchers to achieve position control of magnetic microrobots [42]. Due to the highly negligible mass of the microrobots, their controllers are usually designed with the assumption that the inertia force in the dynamic equation can be ignored. However, due to the nature of the FMs, their mass cannot be ignored. So, the desired force ($F_{des}$) is derived by considering two feed forward terms ($C\,\dot{x}_d + m\,\ddot{x}_d$) in the error-based controller (PI) which is written as Eq (14).

$$F_{des} = k_p\,(x - x_d) + k_I \int (x - x_d)\,dt + C\,\dot{x}_d + m\,\ddot{x}_d \tag{14}$$

The coefficient $k_p$ and $k_I$ correspond to the proportional and integral terms of the controller (the proof of stability is provided in Supplementary Note 2). This PI controller is applied to all three mentioned setups. In order to compare the functionality and energy consumption of these three setups, the same desired trajectory for the FM is considered for all of them, as follows:

$$x_d = 3\,\sin(\frac{2\pi}{50}\,t) \tag{15}$$

where $t$ is time in seconds and $x_d$ is the desired position of the FM from center of the workspace in centimeters and the results for each setup are represented (comparison among setups in Supplementary Note 3).

### 3.1. Experimental results

#### 3.1.1. Setup1: Position control of the FM using the varying Helmholtz coils current

A visual representation of the experimental results for Setup 1 is provided in Figure 5(a) which includes a fixed permanent magnet. Moreover, the desired position of the FM is also shown in the experimental setup and the placement of the FM in the workspace. Figure 5(b) shows the position-time diagram of the FM derived from the Setup1 is displayed with a solid line which is compared with the desired position obtained from Eq (15), shown by the dashed line. The mean absolute error (MAE) between the desired position and actual position of the FM is 0.47 mm. In this approach, the permanent magnet remains stationary while the current flowing through the Helmholtz coils is manipulated to apply both adjusted repulsive and attractive magnetic force to the FM. The desired force is calculated from the controller using Eq (14), leading to the determination of the desired magnetic flux density through Eq (8). The magnitude and direction of the Helmholtz coil current are determined by Eq (13).

Figure 5 (c) illustrates the current of the Helmholtz coils ($I$) versus time during the position control. As shown, the Helmholtz coils' current fluctuates between -1 and +1.2 A to control the FM's position. By comparing Figure 5 (b) and Figure 5 (c), it can be seen that at $t = 0\ s$, the desired position diverges from the permanent magnet, prompting the controller to apply a repulsive magnetic force through a positive current in the Helmholtz coils. Around $t = 10\ s$, before the desired position reaches the pick point at $t = 12.5\ s$, the controller starts to reduce and reverse the current in the Helmholtz coils, leading to switch the magnetic repulsive force to an attractive force. This phenomenon is repeated for the next pick at $t = 27\ s$, where the current of the Helmholtz coils changes before the pick point at $t = 37.5\ s$. From $t = 37.5\ s$ to $50\ s$, the current increases to strengthen the repulsive magnetic force. It is worth mentioning that the magnetic flux density inducing the FM is the summation of the permanent magnet and Helmholtz coils' magnetic flux densities, as discussed in Eq (8). Therefore, a slight positive current in the Helmholtz coils does not necessarily mean that a magnetic repulsive force is being applied. Instead, it is the direction of the summation of the magnetic flux densities ($B$) that matters.

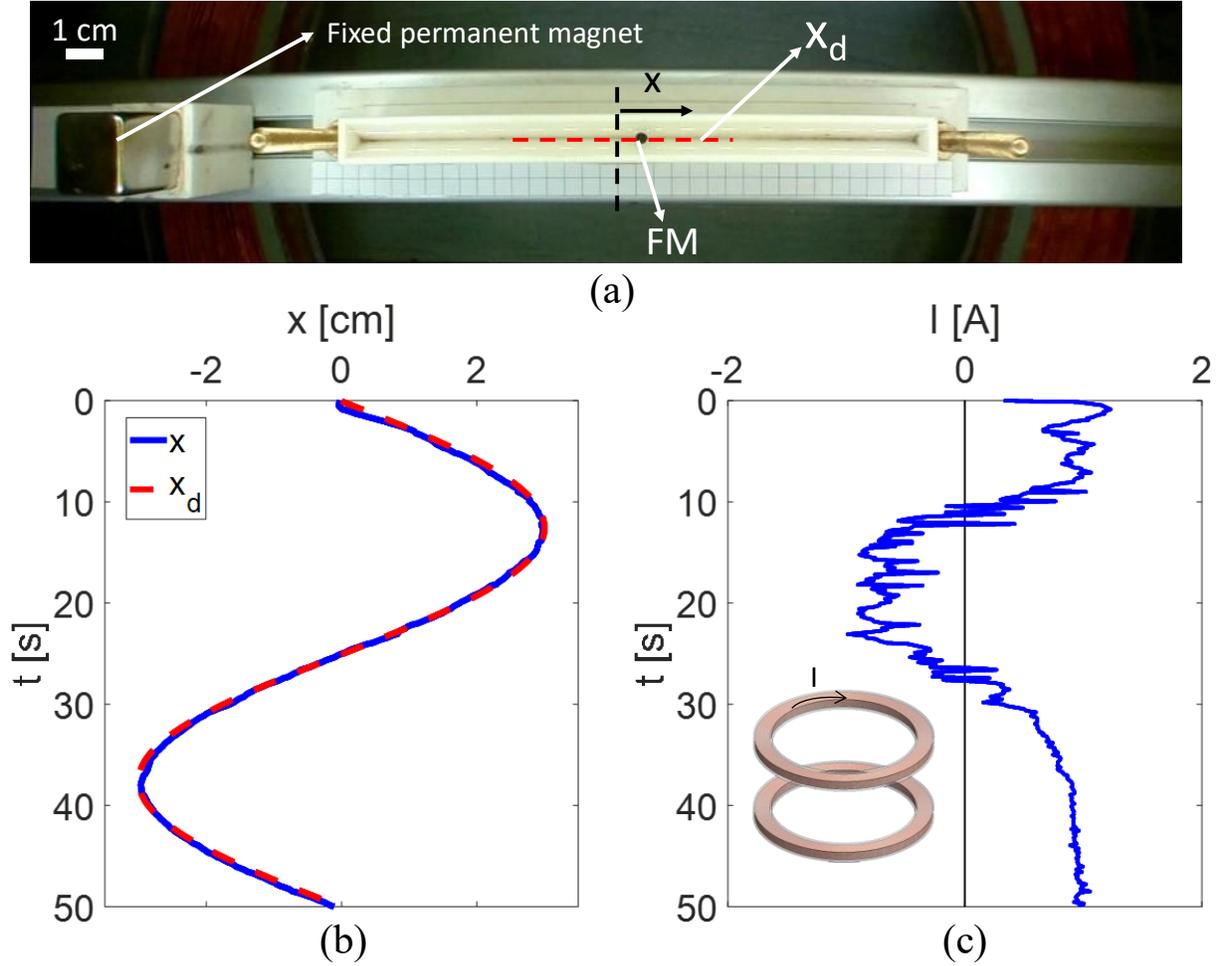

*Figure 5: (a) A snapshot of Setup 1 from the camera view, (b) diagram shows the desired and actual positions of the FM during an experiment, controlled by varying Helmholtz coils current, (c) diagram of the Helmholtz coils current versus time*

### *3.1.2. Setup2: Position control of the FM using one rotating permanent magnet*

In this setup, the stationary permanent magnet is replaced with a rotating one. On the other hand, the current in the Helmholtz coils remains fixed. To ensure a fair comparison, the size of the permanent magnet in Setup 1 and Setup 2 is kept identical. Figure 6(a) presents a snapshot of the experimental setup including a rotating permanent magnet instead of a fixed one. Figure 6(b) illustrates the position-time diagram of the FM and the desired position. The MAE of this position control is 0.94 mm. The FM position is employed to calculate the desired force from Eq (14). Subsequently, the desired angle of the permanent magnet is obtained using Eq (10).

The rotation of the permanent magnet, which applies both attractive and repulsive magnetic forces, is shown in Figure 6 (c). As illustrated, the permanent magnet angle primarily ranges from 30 to 150 degrees, reflecting its role in generating both types of forces. At t = 0 s, the permanent magnet angle is approximately 0 degree, where it exerts a repulsive magnetic force. Generally, the direction of the magnetic force aligns with the direction of the moving FM following the desired trajectory. Specifically, when the desired position diverges from the permanent magnet, the controller sets the magnet's angle between 0 and 90 degrees to apply a repulsive force. Conversely, when the desired position approaches the permanent magnet, an attractive force is applied by adjusting the angle to between 90 and 180 degrees. However, just before the pick times at t = 12.5 s and 37.5 s, the magnitude and direction of the magnetic forces are reduced and reversed to control the inertia of the FM.

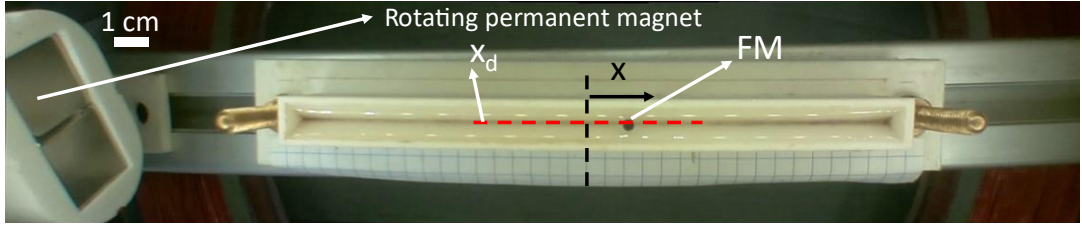

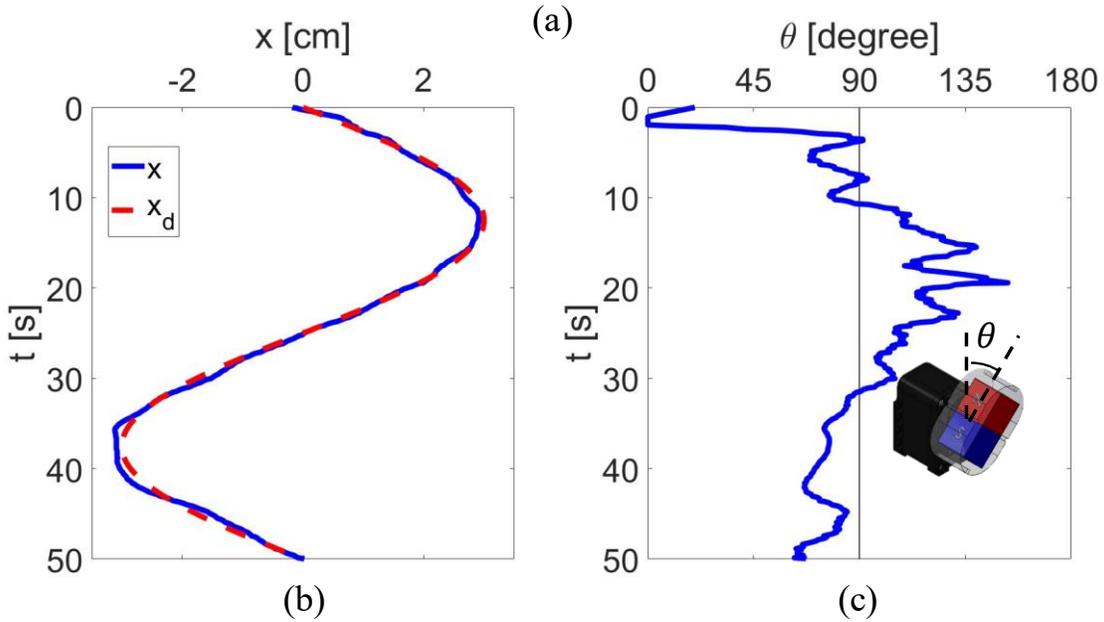

*Figure 6: (a) A snapshot of Setup 2 from the camera view, (b) diagram of the actual FM position and the desired position versus time are depicted, (c) diagram of the permanent magnet angle versus time*

Utilizing a single rotating permanent magnet involves employing both attractive and repulsive forces (Figure 2(b)) to control the position of the FM. However, by substituting two rotating permanent magnets on each side of the workspace, only repulsive magnetic forces can be applied to the FM from two directions (Figure 3 (b)).

### 3.1.3. Setup 3: Position control of FM using two rotating permanent magnets

Figure 7(a) depicts a snapshot of the experimental setup benefit from two rotating permanent magnets. Figure 7(b) depicts the position-time diagram of the FM as it follows the desired trajectory ($x_d$) outlined in Eq (15). The controller computes the desired force based on the error described in Eq (14). Subsequently, these desired forces are transformed into angles of the permanent magnets using Eq (12). Notably, in this setup, the manipulation of the FM relies solely on the repulsive magnetic force. This ensures that there is always a stable equilibrium point in the magnetic force applied to the FM. In other words, it means that both permanent magnets push the FM to get away from the permanent magnet and control the position of the FM.

Figure 7 (c) shows the diagram of the permanent magnets' angles during position control. As illustrated, the angle of the permanent magnets fluctuates between 0 and 90 degrees, ensuring that only repulsive magnetic forces are applied. In general, when the direction of movement in the desired position is to the right side, permanent magnet 1 is fixed at $\theta_1 = 0°$ to apply the maximum repulsive magnetic force to the right, while the angle of permanent magnet 2 is adjusted to control the resultant magnetic force (see Eq (11)). Conversely, when the direction of the movement in the desired position is to the left side the permanent magnet number 2 is fixed at $\theta_2 = 0°$, and the angle of permanent magnet number 1 is adjusted accordingly. However, between $t = 15\ s$ to $20\ s$, when the direction movement in the desired position is to left, the permanent magnet number 2 is adjusted to approximately 50 degrees, to apply desired force to the FM, because the repulsive magnetic force is high at this range of FM distance and should be controlled to apply the desired force. Conversely, this phenomenon is repeated in the time interval between t = 37.5 s and 45s.

To illustrate this equilibrium point, Figure 7 (d) shows the contour of the absolute resultant magnetic force in the workspace during position control. As depicted, the movement of these equilibrium points leads to a stable magnetic force being applied to the FM, making it robust to environmental disturbances. It is important to note that, due to the use of a single magnetic force setup in both Setup 1 and Setup 2, there are no stable equilibrium points.

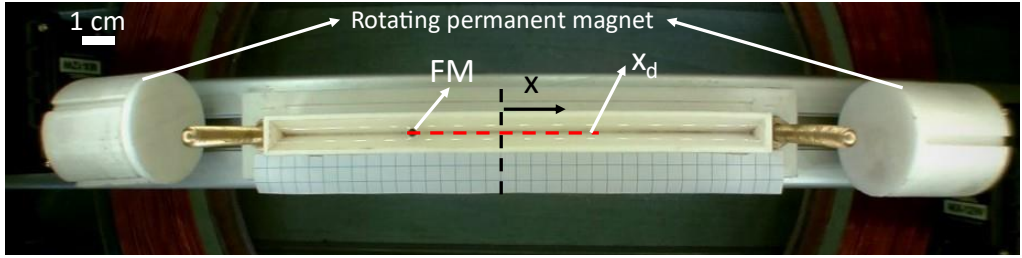

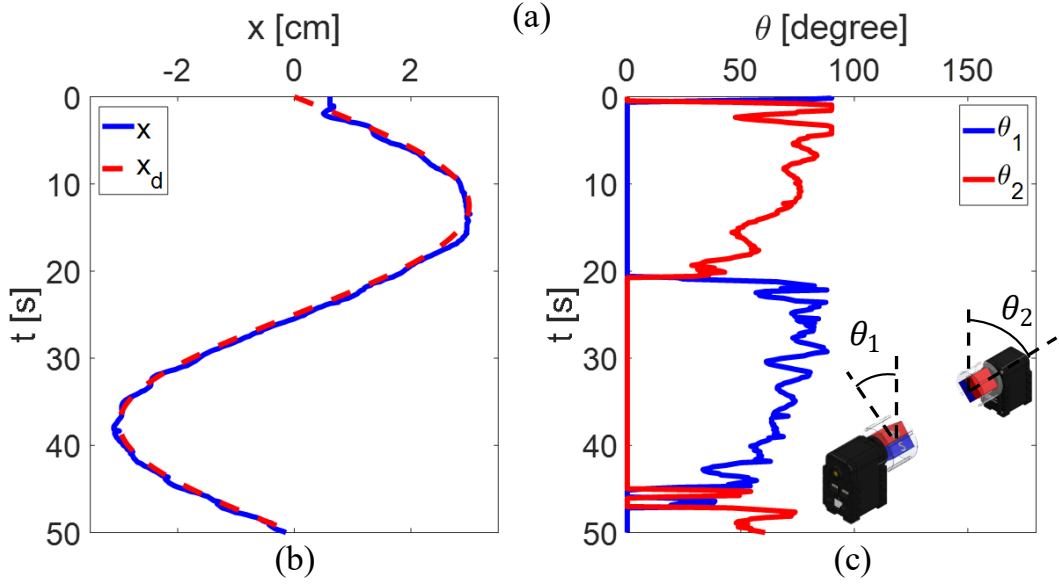

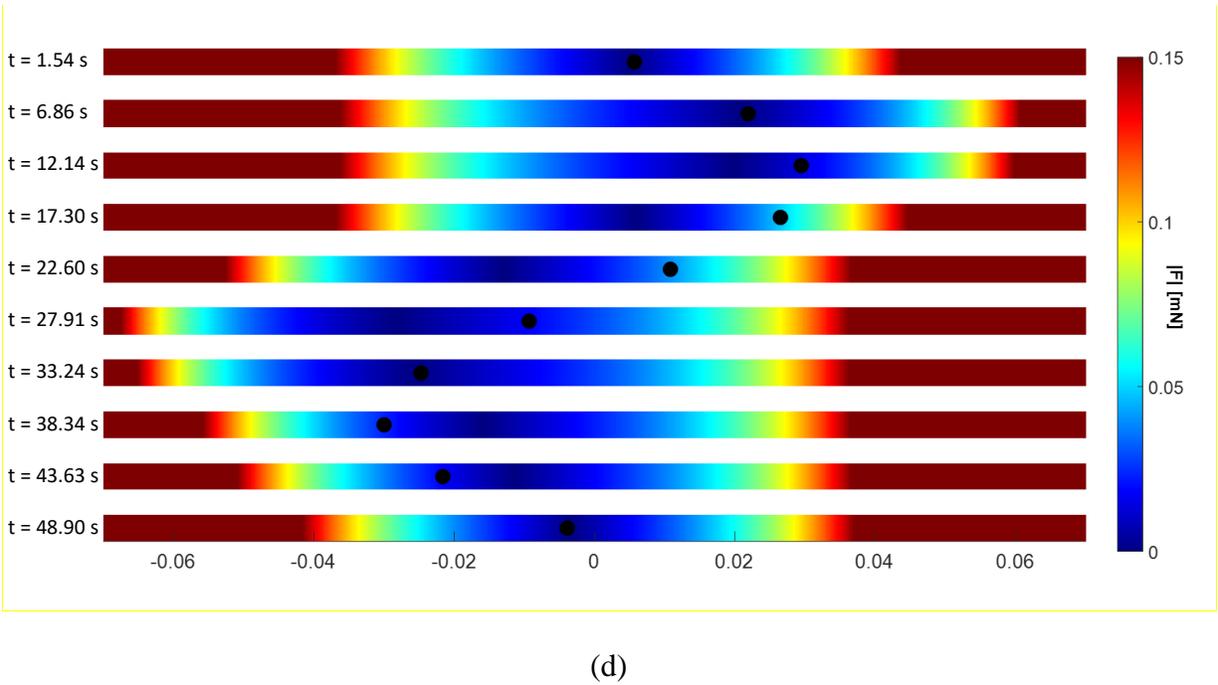

Figure 7: (a) A snapshot from the experimental setup for an FM following desired trajectory, this setup includes two rotating permanent magnets in each side of the workspace and the Helmholtz coils with the fixed current, (b) the diagram illustrates position-time of the FM, the solid lines represent the actual FM position, and the dotted lines show the desired paths, (c)

*diagram of the permanent magnets angles versus time (d) Contour of the absolute resultant magnetic force which is exerted to the FM.*

### 3.2. Position control of two FMs at the same time

The precise manipulation of multiple FMs presents a significant challenge. For the position control of two FMs, it is necessary to apply two distinct magnetic forces. In the proposed setups, Setups 1 and 2 can exert two forces on the FMs, but these forces are not independent. Conversely, Setup 3 benefits from two rotating permanent magnets, providing it with two degrees of freedom. Initially, in Setup 3, the angles of the permanent magnets were configured to apply only repulsive magnetic forces, as shown in Eq (10). However, because this setup uses two rotating permanent magnets, it can apply two distinct magnetic forces. This capability allows for the precise manipulation of two FMs. By concatenating the magnetic forces of these FMs in the form of a matrix, Eq (16) is achieved:

$$\boldsymbol{F_m} = \begin{bmatrix} F_{m1} \\ F_{m2} \end{bmatrix} = A \begin{bmatrix} \frac{1}{r_{11}^4} & -\frac{1}{r_{12}^4} \\ \frac{1}{r_{21}^4} & -\frac{1}{r_{22}^4} \end{bmatrix} \begin{bmatrix} \cos\theta_1 \\ \cos\theta_2 \end{bmatrix} \quad (16)$$

Where in $r_{ij}$, i is the FM ID number and j is the permanent magnets ID number. By rewriting the Eq (16), it can be written as Eq (17).

$$\boldsymbol{F_m} = [R]\,\boldsymbol{C} \quad (17)$$

where $\boldsymbol{C}$ is a vector representing cosine of the permanent magnets angle, and $\boldsymbol{R}$ is a 2×2 matrix that related to the position of the FMs relative to the permanent magnets. In order to evaluate the desired angle of the permanent magnets, Eq (18) is written as follows:

$$\boldsymbol{C} = [R]^{-1}\boldsymbol{F_{des}} \quad (18)$$

The system has two degrees of freedom when both magnetic repulsive and attractive forces are employed, resulting in $C$ values ranging between +1 and -1.

To demonstrate this functionality, two distinct trajectories were chosen that differ in domain, frequency, and center of oscillation. Figure 8 illustrates a snapshot of the experimental setup, showing two FMs following these distinct trajectories. The desired trajectories ($x_d$) for FM number 1 and number 2 are:

$$x_d^1 = 3 + 1.5 \sin\left(\frac{2\pi}{50}t\right), \qquad x_d^2 = -2.5 + \cos\left(\frac{2\pi}{100}t\right) \tag{19}$$

As shown, the current positions of the two FMs follow the desired trajectories, demonstrating the proof of concept for our proposed method.

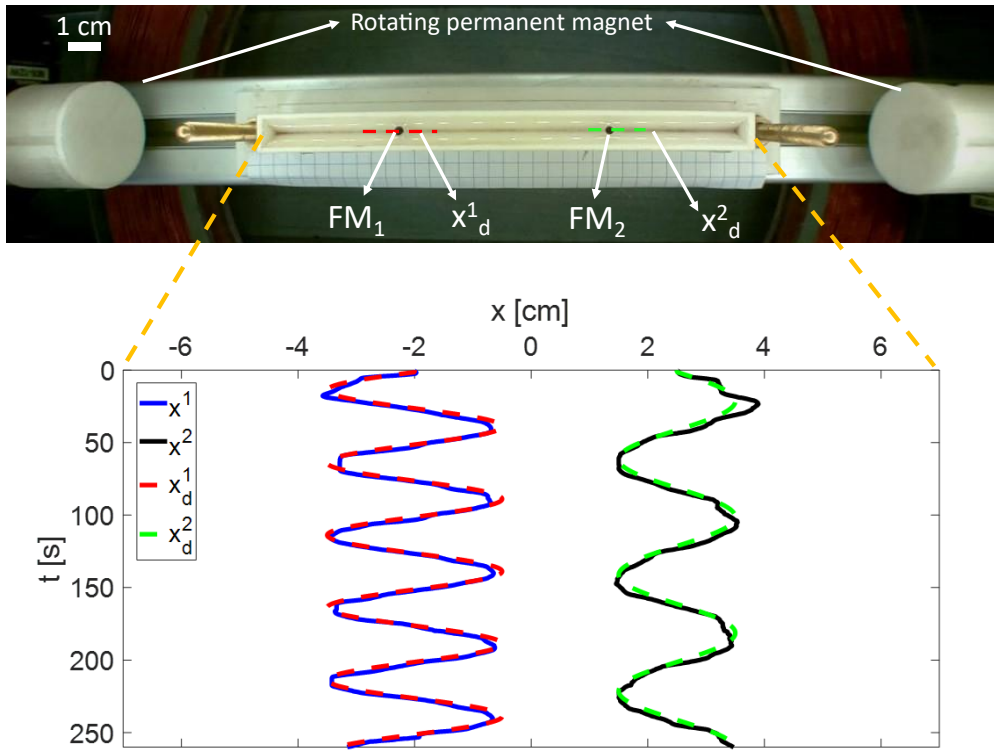

Figure 8: A snapshot of the experimental setup for controlling the positions of two FMs simultaneously. The FMs position versus time diagram shows the desired trajectories and the actual FM positions

## 3.3. Application of position control

### 3.3.1. Reciprocating motion control of FM (mixing and position control simultaneously)

As previously mentioned, Setup 1 benefits from an electrical component to adjust the magnetic flux density. Additionally, Figure S1 demonstrates that this setup operates more efficiently at higher frequencies of reciprocal motion. To leverage this capability, we propose an application involving high-frequency positioning. The aim is to propose a mixing procedure by controllably actuating the FM at a high frequency while it approaches or diverges from the permanent magnet.

The capability of Setup 1 to handle higher frequency variations in the desired trajectory is demonstrated in this section. The FM is controlled using a trajectory that includes higher frequencies, showcasing its ability for reciprocal motion and simultaneous transport. Figure 9 illustrates the two controlled positions of the FM with two different desired trajectories. In the first trajectory, the FM starts fluctuating near the permanent magnet and then controllably diverges from it while maintaining constant reciprocal motion. In the second trajectory, the FM begins fluctuating at a greater distance from the permanent magnet and then controllably approaches it.

In Figure 9(a), the desired position of the FM is given by $x_d = 0.1\, t \sin\left(\frac{2\pi}{6}t\right)$ from $t = 0\, s$ to $t = 20\, s$, which then changes to $x_d = 3 \sin\left(\frac{2\pi}{6}t\right)$ after $t = 20\, s$. During $t = 0\, s$ to $t = 20\, s$ the FM undergoes reciprocal motion while moving away from the permanent magnet and eventually fluctuates around $x = 2\, cm$ after 20 s. In Figure 9(b), the desired trajectory is $x_d = (2 - 0.1\, t) \sin\left(\frac{2\pi}{5}t\right)$ from $t = 0\, s$ to $t = 30\, s$, and then followed by $x_d = -1 \sin\left(\frac{2\pi}{5}t\right)$. The FM initiates back-and-forth movement while approaching the permanent magnet until $t = 30\, s$. This setup illustrates significant capability as a platform for controllable mixing within the FM, showcasing its versatility in handling complex trajectories and high-frequency variations.

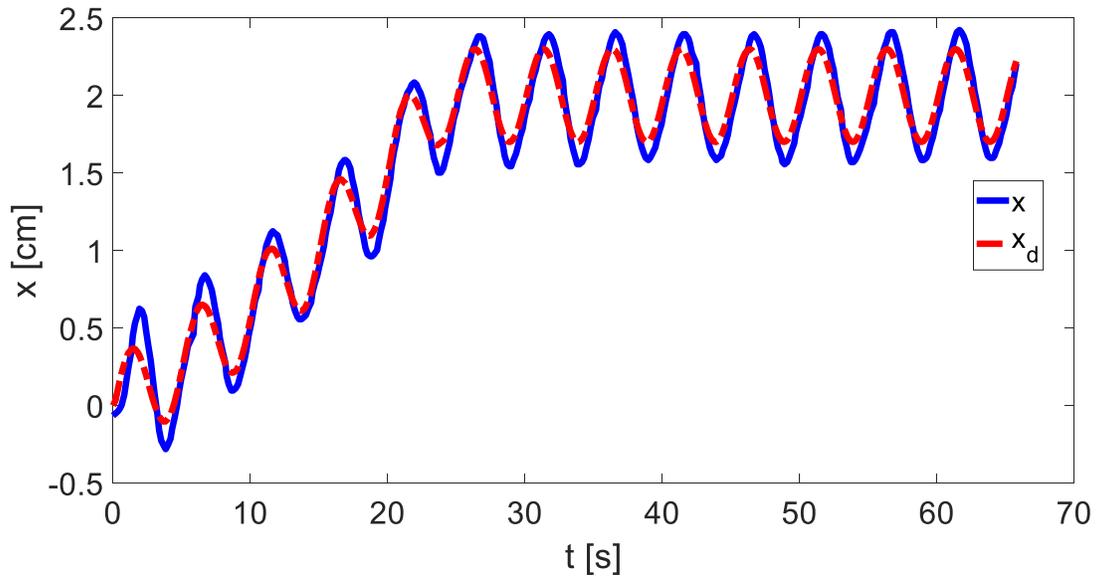

(a)

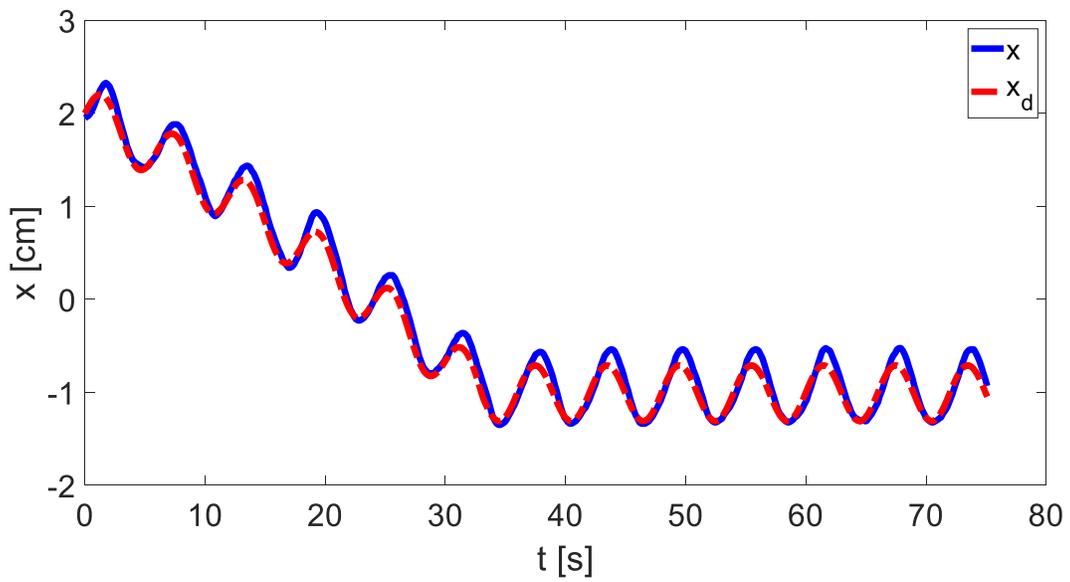

(b)

*Figure 9: The desired and actual FM positions versus time are shown. The desired frequency of reciprocal motion in these tests are higher relative to those in previous sections with the aim of mixing process. a) the oscillation of the FM starts from 0 cm, and the center point of this reciprocal motion increases constantly until t = 20s, at which point it exhibits sinusoidal motion with a constant center line at 2 cm. b) the FM moves closer to the permanent magnet while undergoing back-and-forth motion. After t = 30 s, the FM achieves the oscillation with a constant center line at -1 cm.*

### *3.3.2. Manipulation of non-magnetic liquid marble using the FM*

Precise position control of FMs enables the precise manipulation of non-magnetic components, showcasing one of its specific capabilities. In this section, we introduce a novel method for positioning non-magnetic liquid marbles using the FMs. In this method, the FM carries a non-magnetic liquid marble and by position controlling the FM, the precise positioning of non-magnetic marble could be achieved. Although all three proposed setups can achieve this task, we utilize Setup 3 to illustrate this concept as an example. Figure 10(a) illustrates an FM and a non-magnetic liquid marble attached together and placed on the water surface. As can be seen, the non-magnetic marble appears white, synthesized using Poly Tetra Fluoro Ethylene (PTFE) powder (hydrophobic nanoparticles) around a water droplet, while the FM next to it, is black.

Figure 10(b) depicts the manipulation of the non-magnetic liquid marble by the FM using Setup 3 which consists of a constant current in the Helmholtz coil and two rotating permanent magnets. At $t = 0\ s$, the FM is positioned on the left side of the workspace, and non-magnetic liquid marble is on the right side. At $t = 10\ s$, the FM is moving towards the non-magnetic liquid marble considering a desired trajectory of $x_d = 2\sin(\frac{2\pi}{80}t)$. Due to the capillary forces and concavity of the water surface, they collapse together and remain attached as shown at t = 18 s. By applying the magnetic force to the FM, the non-magnetic marble is also manipulated.

At t = 35 s, pulls the non-magnetic marble, with the capillary force between them being dominant compared to friction and inertia forces. As a result, the non-magnetic marble moves with the same trajectory as that of the FM. The FM even has the ability to pushes the non-magnetic liquid marble.

During the time interval between t = 60 s and t = 70 s the FM push the non-magnetic liquid marble based on the desired position of the FM. Finally, to dispense these marbles, the maximum magnetic force is applied to the FM to move it to the left. In Setup 3, this is achieved by adjusting the permanent magnet angle on the left to $\theta = 180°$ and the permanent magnet angle on the right to $\theta = 0°$.

Figure 10(c) plots the desired position and the actual position of the FM over time. Generally, even with a non-magnetic marble attached, the FM is robust enough to carry it controllably along the desired trajectory. During the grabbing moments at $t = 8\ s$ and $t = 18\ s$, the capillary forces between the FM and the non-magnetic liquid marble act as a disturbance in the position control of the FM. However, by applying the desired force to the FM, the actuators successfully regain control of its position, allowing it to follow the trajectory accurately.

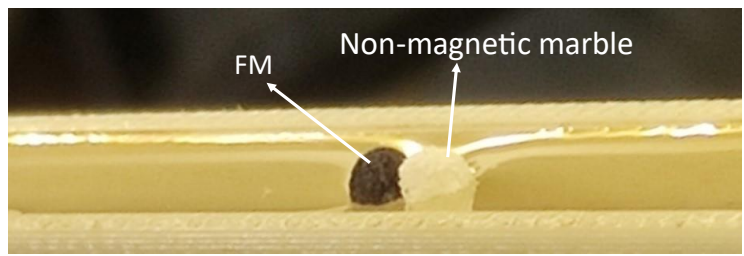

(a)

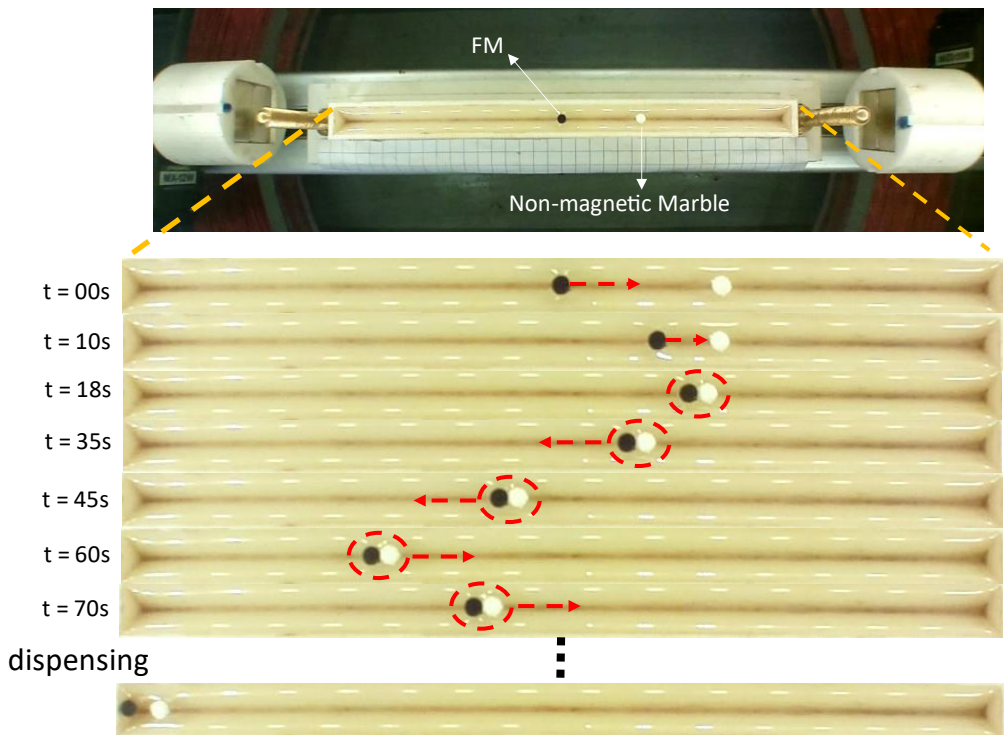

(b)

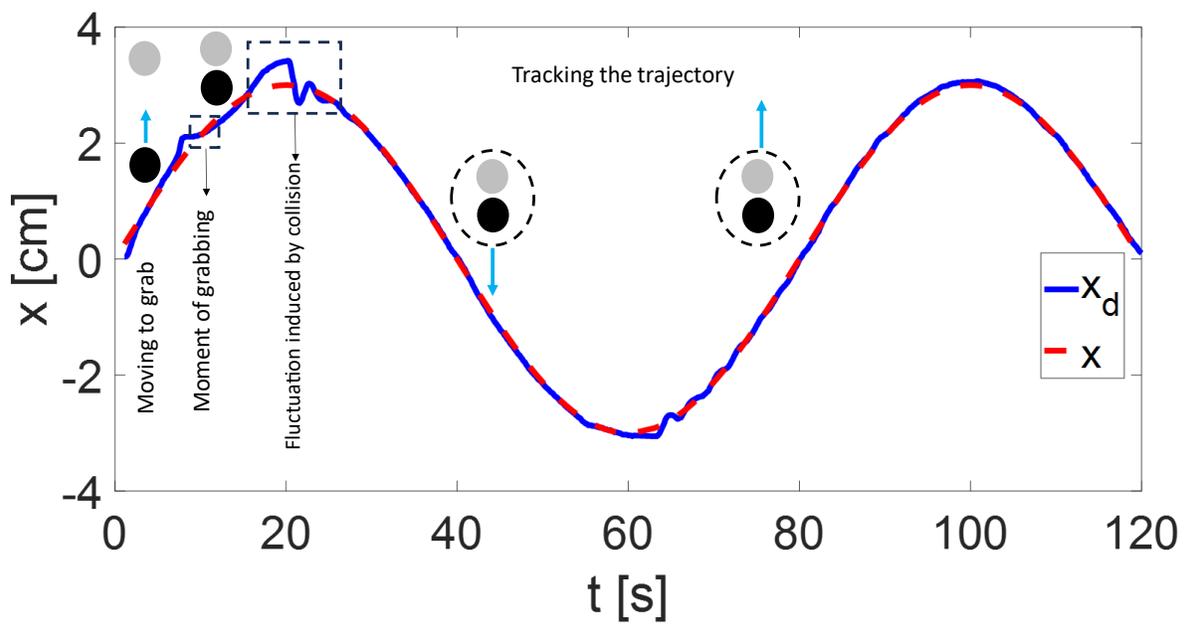

(c)

*Figure 10: Transportation of the non-magnetic liquid marble using the FM, (a) a real snapshot of non-magnetic liquid marble is placed beside the FM, (b) the procedure of manipulating the non-magnetic marble, at first step the FM is moving toward the non-magnetic marble. Then the FM and non-magnetic marble adhere to each other as a result of capillary forces. When they are attached to each other the FM can pull or push the non-magnetic marble, (c) position versus time diagram of the FM which is tracking the desired trajectory. At t = 18s, the FM and non-magnetic marble are attached. Because of capillary forces, it acts like a disturbance. But the controller is robust to this disturbance and can track the desired trajectory to the end.*

### *3.3.3. Sample extraction*

Extraction of liquid from the core of the marbles presents challenges due to their small size and the solid powder surrounding them. In this section, for the first time, a technique employing position control of the FM is proposed for a sample extraction scheme from the core of the marble which is a liquid. To this aim, a microneedle is used to penetrate into marbles, utilizing their high capillary characteristics to extract liquid from the core of the liquid marble. This microneedle is fabricated from a pre-pulled borosilicate micropipette, with a tip diameter ranging from 20 to 50 micrometers. It is produced using a P-97 Sutter Instrument device, which pulls capillary glass with an outer diameter of 1.6 mm. By using a FM that moves through a microneedle controllability then applying the extraction of the ferrofluid used as the core of the marble, conducted. The powder used for generating these FM is PTFE due to the larger size of these powders, the chance of microneedle blockage during the sample extraction procedure reduces. The microneedle is made from Borosilicate glass.

Figure 11 (a) briefly illustrates the concept of the procedure. Initially, the FM position is controlled using the magnetic field in Setup3 (similar processes can be conducted using other mentioned setups, and here Setup3 is used as an example). When the FM contacts the tip of the microneedle, the permanent magnet on the right applies the maximum attractive force by setting its angle to $\theta = 180°$, while the permanent magnet on the left applies the maximum repulsive force ($\theta = 0°$). Consequently, the resultant force is maximized to the right. These forces are applied until the microneedle penetrates the FM. Capillary forces between the ferrofluid inside the FM and the microneedle trigger the extraction of the sample. This process continues until all the ferrofluid is extracted into the microneedle.

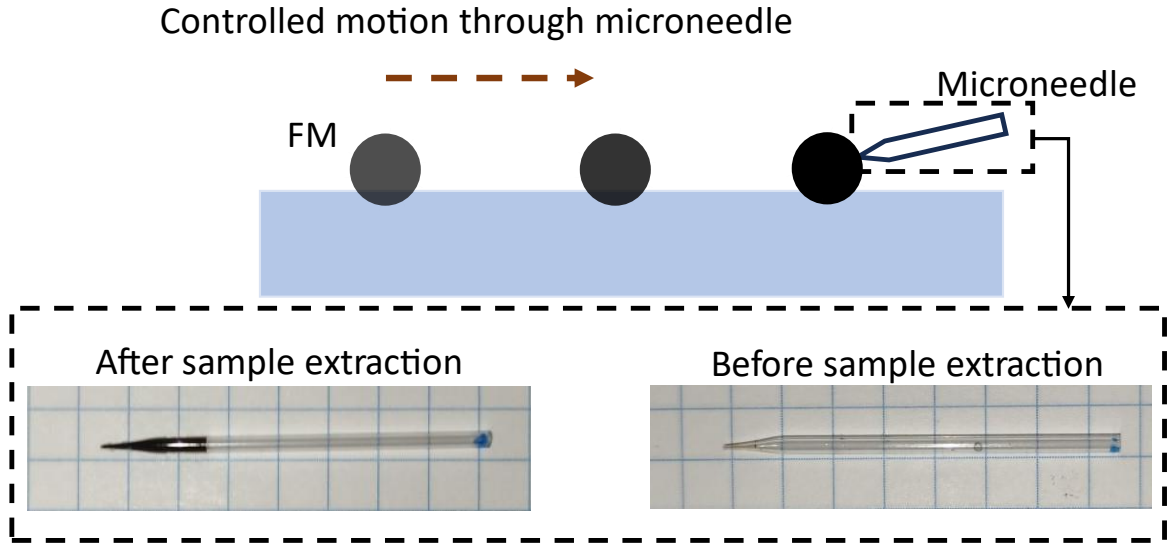

(a)

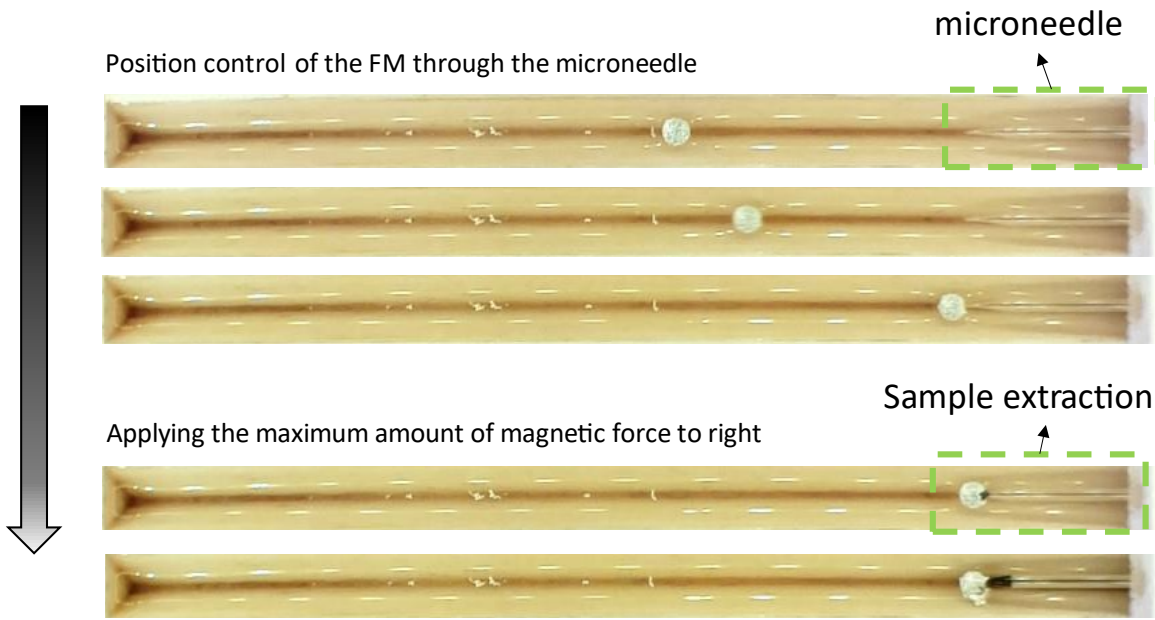

(b)

*Figure 11 (a) Schematic of the process. (b) Snapshots of the experiment: The FM moves controllably towards the microneedle. Upon contact, the microneedle touches the FM, and the permanent magnet on the right attracts the FM with maximum*

*magnetic force until the needle penetrates the FM. Capillary forces between the FM and the needle then cause extraction of the ferrofluid from inside the marble.*

## 4. Conclusion

The precise positioning of the FMs has been always challenging in digital magnetic microfluidics. In this study, a closed-loop control is proposed for the manipulation of the FM using three different setups benefiting from repulsive and attractive forces. To generate these forces, the Helmholtz coils for producing the uniform magnetic field and the permanent magnets are utilized. At the first setup, the permanent magnet is fixed and the Helmholtz coils current is controlled. In the second one, the Helmholtz coils current is fixed, but the setup uses a rotating permanent magnet. In the third one, the current is fixed too; But uses two rotating permanent magnets to apply just magnetic repulsive force. These setups are compared in terms of energy consumption and their ability to control FM position in different trajectories. It is proven that Setup1 consumes less energy than other setups and controls the position of the FM more precisely. On the other hand, Setup3 has the ability to control the position of two marbles simultaneously and makes a stable equilibrium point in position control of the FMs. Moreover, three applications of position control of the FM are proposed including, mixing and moving the FMs simultaneously, carrying non-magnetic liquid marbles, and sample extraction from the FM are proposed.

Future studies can focus more on the proposed applications. For instance, the sample extraction phenomenon, which is briefly introduced in this paper, can be investigated in more detail. Carrying non-magnetic marbles also has several applications in biology and chemistry. From a control theory perspective, employing a more advanced controller could provide increased robustness to disturbances.

**Supporting information**

Supporting movie 1: Position control of the FM using Setup1 in Figure 5

Supporting movie 2: Position control of the FM using Setup2 in Figure 6

Supporting movie 3: Position control of the FM using Setup3 in Figure 7

Supporting Note 1: Experimental Setup

Supplementary Note 2: Designing controller

Supplementary Note 3: Comparison of setups

## 5. Reference

[1]     Y. Luo, Z. Zheng, X. Zheng, Y. Li, Z. Che, J. Fang, L. Xi, N.T. Nguyen, C. Song, Model-based feedback control for on-demand droplet dispensing system with precise real-time phase imaging, Sensors Actuators B Chem. 365 (2022) 131936. https://doi.org/10.1016/j.snb.2022.131936.

[2]     U. Roshan, Y. Dai, A.S. Yadav, S. Hettiarachchi, A. Mudugamuwa, J. Zhang, N.T. Nguyen, Flexible droplet microfluidic devices for tuneable droplet generation, Sensors Actuators B Chem. 422 (2025) 136617. https://doi.org/10.1016/j.snb.2024.136617.

[3]     Z. Jiang, H. Shi, X. Tang, J. Qin, Recent advances in droplet microfluidics for single-cell analysis, TrAC - Trends Anal. Chem. 159 (2023) 116932. https://doi.org/10.1016/j.trac.2023.116932.

[4]   J. Wang, Y. Li, X. Wang, J. Wang, H. Tian, P. Zhao, Y. Tian, Y. Gu, L. Wang, C. Wang, Droplet microfluidics for the production of microparticles and nanoparticles, Micromachines. 8 (2017) 1–23. https://doi.org/10.3390/mi8010022.

[5]   T.N.D. Trinh, H.D.K. Do, N.N. Nam, T.T. Dan, K.T.L. Trinh, N.Y. Lee, Droplet-Based Microfluidics: Applications in Pharmaceuticals, Pharmaceuticals. 16 (2023). https://doi.org/10.3390/ph16070937.

[6]   K. Schroen, C. Berton-Carabin, D. Renard, M. Marquis, A. Boire, R. Cochereau, C. Amine, S. Marze, Droplet microfluidics for food and nutrition applications, Micromachines. 12 (2021). https://doi.org/10.3390/mi12080863.

[7]   K. Choi, A.H.C. Ng, R. Fobel, A.R. Wheeler, Digital Microfluidics, Http://Dx.Doi.Org/10.1146/Annurev-Anchem-062011-143028. 5 (2012) 413–440. https://doi.org/10.1146/ANNUREV-ANCHEM-062011-143028.

[8]   B. Wang, K.F. Chan, F. Ji, Q. Wang, P.W.Y. Chiu, Z. Guo, L. Zhang, On-Demand Coalescence and Splitting of Liquid Marbles and Their Bioapplications, Adv. Sci. 6 (2019). https://doi.org/10.1002/advs.201802033.

[9]   V. Sivan, S. Tang, A.P. O'Mullane, P. Petersen, N. Eshtiaghi, K. Kalantar-zadeh, A. Mitchell, Liquid Metal Marbles, Adv. Funct. Mater. 23 (2013) 144–152. https://doi.org/10.1002/adfm.201200837.

[10]  S. Mahmoudi Salehabad, S. Azizian, S. Fujii, Shape-Designable Liquid Marbles Stabilized by Gel Layer, Langmuir. 35 (2019) 8950–8960. https://doi.org/10.1021/acs.langmuir.9b01473.

[11]  N.K. Nguyen, P. Singha, Y. Dai, K. Rajan Sreejith, D.T. Tran, H.P. Phan, N.T. Nguyen, C.H. Ooi, Controllable high-performance liquid marble micromixer, Lab Chip. 22 (2022) 1508–1518. https://doi.org/10.1039/d2lc00017b.

[12]  N. Kavokine, M. Anyfantakis, M. Morel, S. Rudiuk, T. Bickel, D. Baigl, Light-Driven Transport of a Liquid Marble with and against Surface Flows, Angew. Chemie Int. Ed. 55 (2016) 11183–11187. https://doi.org/10.1002/anie.201603639.

[13]  P. Aussillous, D. Quéré, Properties of liquid marbles, Proc. R. Soc. A Math. Phys. Eng. Sci. 462 (2006) 973–999. https://doi.org/10.1098/rspa.2005.1581.

[14]  P. AUSSILLOUS, D. QUÉRÉ, Shapes of rolling liquid drops, J. Fluid Mech. 512 (2004). https://doi.org/10.1017/S0022112004009747.

[15]  E. Bormashenko, Liquid marbles: Properties and applications, Curr. Opin. Colloid Interface Sci. 16 (2011) 266–271. https://doi.org/10.1016/j.cocis.2010.12.002.

[16]  K.R. Sreejith, L. Gorgannezhad, J. Jin, C.H. Ooi, H. Stratton, D.V. Dao, N.-T. Nguyen, Liquid marbles as biochemical reactors for the polymerase chain reaction, Lab Chip. 19 (2019) 3220–3227. https://doi.org/10.1039/C9LC00676A.

[17]  M. Paven, H. Mayama, T. Sekido, H.-J. Butt, Y. Nakamura, S. Fujii, Light-Driven Delivery and Release of Materials Using Liquid Marbles, Adv. Funct. Mater. 26 (2016) 3199–3206. https://doi.org/10.1002/adfm.201600034.

[18]  M.I. Newton, D.L. Herbertson, S.J. Elliott, N.J. Shirtcliffe, G. McHale, Electrowetting of liquid marbles, J. Phys. D. Appl. Phys. 40 (2007) 20–24. https://doi.org/10.1088/0022-3727/40/1/S04.

[19] M.G. Pollack, A.D. Shenderov, R.B. Fair, Electrowetting-based actuation of droplets for integrated microfluidicsElectronic supplementary information (ESI) available: six videos showing droplet flow, droplet dispensing and electrowetting. See http://www.rsc.org/suppdata/lc/b1/b110474h/, Lab Chip. 2 (2002) 96. https://doi.org/10.1039/b110474h.

[20] M. Mohammadrashidi, M.A. Bijarchi, M.B. Shafii, M. Taghipoor, Experimental and Theoretical Investigation on the Dynamic Response of Ferrofluid Liquid Marbles to Steady and Pulsating Magnetic Fields, Langmuir. (2023). https://doi.org/10.1021/acs.langmuir.2c02811.

[21] D.J. Collins, T. Alan, K. Helmerson, A. Neild, Surface acoustic waves for on-demand production of picoliter droplets and particle encapsulation, Lab Chip. 13 (2013) 3225. https://doi.org/10.1039/c3lc50372k.

[22] Y. Zhang, N.-T. Nguyen, Magnetic digital microfluidics – a review, Lab Chip. 17 (2017) 994–1008. https://doi.org/10.1039/C7LC00025A.

[23] X. Hu, Y. Zhang, J. Yang, K. Xiao, J. Guo, X. Zhang, Magnetic digital microfluidic manipulation with mobile surface energy traps capable of releasable droplet dispensing, Sensors Actuators B Chem. 393 (2023) 134283. https://doi.org/10.1016/j.snb.2023.134283.

[24] R.E. Rosensweig, Ferrohydrodynamics, Courier Corporation, 2013.

[25] R.E. Rosensweig, Magnetic Fluids, Annu. Rev. Fluid Mech. 19 (1987) 437–461. https://doi.org/10.1146/annurev.fl.19.010187.002253.

[26] W. Yu, H. Lin, Y. Wang, X. He, N. Chen, K. Sun, D. Lo, B. Cheng, C. Yeung, J. Tan, D. Di Carlo, S. Emaminejad, A ferrobotic system for automated microfluidic logistics, Sci. Robot. 5 (2020) eaba4411. https://doi.org/10.1126/scirobotics.aba4411.

[27] R. Yang, H. Hou, Y. Wang, L. Fu, Micro-magnetofluidics in microfluidic systems: A review, Sensors Actuators B Chem. 224 (2016) 1–15. https://doi.org/10.1016/j.snb.2015.10.053.

[28] H. Hartshorne, C.J. Backhouse, W.E. Lee, Ferrofluid-based microchip pump and valve, Sensors Actuators B Chem. 99 (2004) 592–600. https://doi.org/10.1016/j.snb.2004.01.016.

[29] M.H. Sarkhosh, M. Yousefi, M.A. Bijarchi, H. Nejat Pishkenari, K. Forghani, Manipulation of ferrofluid marbles and droplets using repulsive force in magnetic digital microfluidics, Sensors Actuators A Phys. 363 (2023) 114733. https://doi.org/10.1016/j.sna.2023.114733.

[30] D. Chakrabarty, S. Dutta, N. Chakraborty, R. Ganguly, Magnetically actuated transport of ferrofluid droplets over micro-coil array on a digital microfluidic platform, Sensors Actuators B Chem. 236 (2016) 367–377. https://doi.org/10.1016/j.snb.2016.06.001.

[31] M. Youesfi, M.H. Sarkhosh, M.A. Bijarchi, H.N. Pishkenari, Investigating the Response of Ferrofluid Droplets in a Uniform Rotating Magnetic Field: Towards Splitting and Merging, in: 2023 Int. Conf. Manip. Autom. Robot. Small Scales, 2023: pp. 1–5.

[32] S. AFKHAMI, A.J. TYLER, Y. RENARDY, M. RENARDY, T.G. St. PIERRE, R.C. WOODWARD, J.S. RIFFLE, Deformation of a hydrophobic ferrofluid droplet suspended in a viscous medium under uniform magnetic fields, J. Fluid Mech. 663 (2010) 358–384. https://doi.org/10.1017/S0022112010003551.

[33] P. Zhao, L. Yan, X. Gao, A programmable ferrofluidic droplet robot, Soft Matter. 46 (2023) 87.

[34] M.A. Bijarchi, M. Dizani, M. Honarmand, M.B. Shafii, Splitting dynamics of ferrofluid droplets inside a microfluidic T-junction using a pulse-width modulated magnetic field in micro-magnetofluidics, Soft Matter. 17 (2021) 1317–1329. https://doi.org/10.1039/D0SM01764G.

[35] X. Fan, M. Sun, L. Sun, H. Xie, Ferrofluid Droplets as Liquid Microrobots with Multiple Deformabilities, Adv. Funct. Mater. 30 (2020) 1–12. https://doi.org/10.1002/adfm.202000138.

[36] M.A. Bijarchi, A. Favakeh, S. Alborzi, M.B. Shafii, Experimental investigation of on-demand ferrofluid droplet generation in microfluidics using a Pulse-Width Modulation magnetic field with proposed correlation, Sensors Actuators B Chem. 329 (2021) 129274. https://doi.org/10.1016/j.snb.2020.129274.

[37] A. Ray, V.B. Varma, P.J. Jayaneel, N.M. Sudharsan, Z.P. Wang, R. V. Ramanujan, On demand manipulation of ferrofluid droplets by magnetic fields, Sensors Actuators B Chem. 242 (2017) 760–768. https://doi.org/10.1016/J.SNB.2016.11.115.

[38] R. Ahmed, M. Ilami, J. Bant, B. Beigzadeh, H. Marvi, A Shapeshifting Ferrofluidic Robot, 8 (2021) 687–698. https://doi.org/10.1089/soro.2019.0184.

[39] H. Lin, W. Yu, K. A. Sabet, M. Bogumil, Y. Zhao, J. Hambalek, S. Lin, S. Chandrasekaran, O. Garner, D. Di Carlo, S. Emaminejad, Ferrobotic swarms enable accessible and adaptable automated viral testing, Nature. 611 (2022) 570–577. https://doi.org/10.1038/s41586-022-05408-3.

[40]  L. Yang, L. Zhang, Motion Control in Magnetic Microrobotics: From Individual and Multiple Robots to Swarms, Annu. Rev. Control. Robot. Auton. Syst. 4 (2021) 509–534. https://doi.org/10.1146/annurev-control-032720-104318.

[41]  R. Khalesi, M. Yousefi, H. Nejat Pishkenari, G. Vossoughi, Robust independent and simultaneous position control of multiple magnetic microrobots by sliding mode controller, Mechatronics. 84 (2022) 102776. https://doi.org/10.1016/J.MECHATRONICS.2022.102776.

[42]  M. Yousefi, H. Nejat Pishkenari, Independent position control of two identical magnetic microrobots in a plane using rotating permanent magnets, J. Micro-Bio Robot. 2021 171. 17 (2021) 59–67. https://doi.org/10.1007/S12213-021-00143-W.

[43]  S.A. Abbasi, A. Ahmed, S. Noh, N.L. Gharamaleki, S. Kim, A.M.M.B. Chowdhury, J. Kim, S. Pané, B.J. Nelson, H. Choi, Autonomous 3D positional control of a magnetic microrobot using reinforcement learning, Nat. Mach. Intell. 6 (2024) 92–105. https://doi.org/10.1038/s42256-023-00779-2.

[44]  X. Fan, X. Dong, A.C. Karacakol, H. Xie, M. Sitti, Reconfigurable multifunctional ferrofluid droplet robots, Proc. Natl. Acad. Sci. U. S. A. 117 (2020) 27916–27926. https://doi.org/10.1073/PNAS.2016388117.

[45]  X. Fan, Y. Jiang, M. Li, Y. Zhang, C. Tian, L. Mao, H. Xie, L. Sun, Z. Yang, M. Sitti, Scale-reconfigurable miniature ferrofluidic robots for negotiating sharply variable spaces, Sci. Adv. 8 (2022). https://doi.org/10.1126/sciadv.abq1677.

[46]  Y. Xue, H. Wang, Y. Zhao, L. Dai, L. Feng, X. Wang, T. Lin, Magnetic liquid marbles: A "precise" miniature reactor, Adv. Mater. 22 (2010) 4814–4818.

https://doi.org/10.1002/adma.201001898.

[47] R. Zhu, M. Liu, Y. Hou, L. Zhang, M. Li, D. Wang, S. Fu, One-Pot Preparation of Fluorine-Free Magnetic Superhydrophobic Particles for Controllable Liquid Marbles and Robust Multifunctional Coatings, ACS Appl. Mater. Interfaces. 12 (2020) 17004–17017. https://doi.org/10.1021/acsami.9b22268.

[48] Q. Wang, N. Xiang, J. Lang, B. Wang, D. Jin, L. Zhang, Reconfigurable Liquid-Bodied Miniature Machines: Magnetic Control and Microrobotic Applications, Adv. Intell. Syst. 6 (2024). https://doi.org/10.1002/aisy.202300108.

[49] E. Bormashenko, R. Pogreb, Y. Bormashenko, A. Musin, T. Stein, New Investigations on Ferrofluidics: Ferrofluidic Marbles and Magnetic-Field-Driven Drops on Superhydrophobic Surfaces, Langmuir. 24 (2008) 12119–12122. https://doi.org/10.1021/la802355y.

[50] H. Dayyani, A. Mohseni, M.A. Bijarchi, Dynamic behavior of floating magnetic liquid marbles under steady and pulse-width-modulated magnetic fields, Lab Chip. (2024). https://doi.org/10.1039/D3LC00578J.

[51] M.K. Khaw, C.H. Ooi, F. Mohd-Yasin, R. Vadivelu, J.S. John, N.-T. Nguyen, Digital microfluidics with a magnetically actuated floating liquid marble, Lab Chip. 16 (2016) 2211–2218. https://doi.org/10.1039/C6LC00378H.

[52] C.H. Ooi, A. Van Nguyen, G.M. Evans, D.V. Dao, N.T. Nguyen, Measuring the Coefficient of Friction of a Small Floating Liquid Marble, Sci. Rep. 6 (2016). https://doi.org/10.1038/srep38346.